\documentclass[12pt]{article}
\usepackage[a4paper,margin=1in]{geometry}
\usepackage{graphicx,amsmath,amssymb,bm}
\usepackage{caption}
\usepackage{subcaption}
\usepackage{authblk}
\usepackage{hyperref}

\def\BibTeX{{\rm B\kern-.05em{\sc i\kern-.025em b}\kern-.08em
    T\kern-.1667em\lower.7ex\hbox{E}\kern-.125emX}}

\usepackage{makecell}
\usepackage{lastpage}
\usepackage{fancyhdr}
\usepackage{lipsum}

\begin{document}

\title{Reconfigurable Auxetic Devices (RADs) with Backlash for Soft Robotic Surface Manipulation}

\author[1]{Jacob Miske*}
\author[2]{Ahyan Maya}
\author[3]{Ahnaf Inkiad}
\author[1]{Jeffrey Lipton}

\affil[1]{Department of Mechanical and Industrial Engineering, Northeastern University, Boston, MA, USA}
\affil[2]{Department of Mathematics, Northeastern University, Boston, MA, USA}
\affil[3]{Khoury College of Computer Sciences, Northeastern University, Boston, MA, USA}
\affil[*]{Corresponding author email: miske.j@northeastern.edu}

\maketitle

\begin{abstract}
    Robotic surfaces traditionally use materials with a positive Poisson's ratio to push and pull on a manipulation interface. Auxetic materials with a negative Poisson's ratio may expand in multiple directions when stretched and enable conformable interfaces. Here we demonstrate reconfigurable auxetic lattices for robotic surface manipulation. Our approach enables shape control through reconfigurable locking or embedded servos that underactuate an auxetic lattice structure. Variable expansion of local lattice areas is enabled by backlash between unit cells. Demonstrations of variable surface conformity are presented with characterization metrics. Experimental results are validated against a simplified model of the system, which uses an activation function to model intercell coupling with backlash. Reconfigurable auxetic structures are shown to achieve manipulation via variable surface contraction and expansion. This structure maintains compliance with backlash in contrast with previous work on auxetics, opening new opportunities in adaptive robotic structures for surface manipulation tasks.
\end{abstract}

\section*{Keywords}
    auxetic, soft robotics, compliance, backlash, dynamic modeling, ReLU

\newpage
\section{Introduction}
Dynamic control over morphing surfaces is a developing area in robotic manipulation. We believe that future robots must be able to manipulate, translate, and sense over surfaces that vary in shape over time. Variable geometry auxetic structures are currently used to produce such robotic surfaces in aerospace manufacturing and biomedical applications, with examples relying on fixed geometric patterns or material compositions that do not adapt during operation \cite{kristensen2015flexible, konakovic2016beyond, Rauf2023Electroadhesive}. These monolithic auxetic structures include unit cells with preset mechanical responses, which results in structures with limited adaptability to varying surface conditions or a reduction in performance when encountering unexpected geometries. If we had auxetic structures that conform to a wide range of surfaces, that may reduce these issues and enable further utility of these adaptive structures.

We now see structures for robotic manipulation that use tunable, gradient, and reconfigurable mechanical response or Poisson's ratios \cite{bertoldi_auxetic_2010, HasanshahiMachines12080527}. And prior applications for auxetic-based robotic systems have demonstrated variable expansion ratios for gripping mechanisms and conformable interfaces \cite{wang2017foam, Liu_roboticsurface_2021, steed_mechatronic_2021}. These studies show the importance of geometric design and material selection, but rely on single material fabrication and 1D transformations between flat and curved during operation between physical states. 
\begin{figure}[!h]
    \centering
    \includegraphics[width=4in]{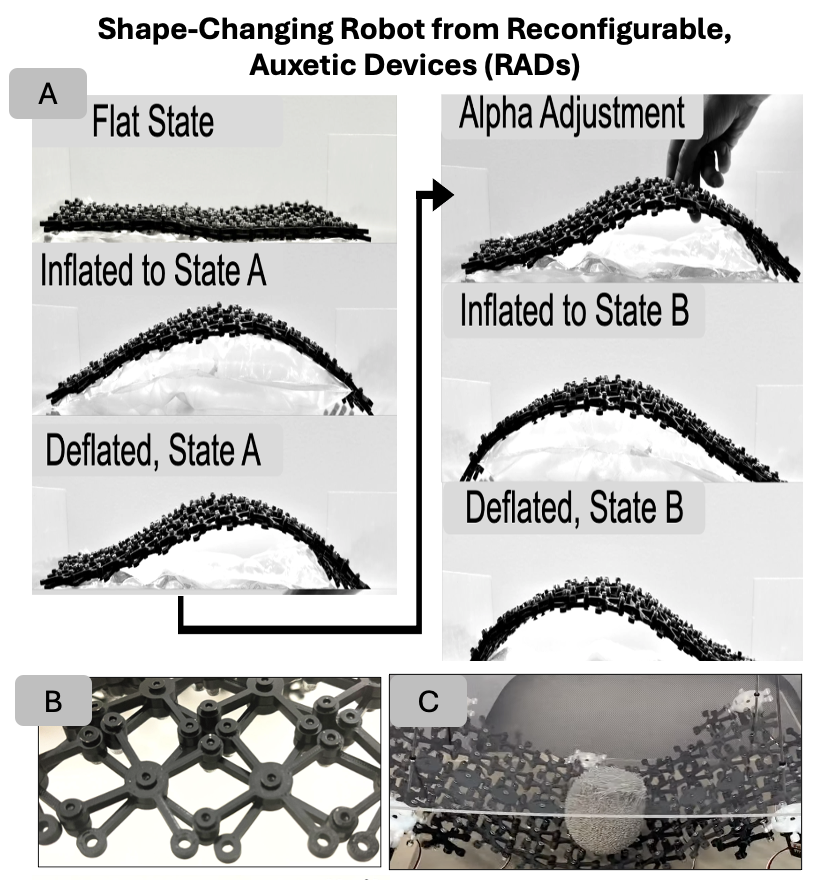}
    \caption{(A) Using a lattice of semi-rigid auxetic cells with significant backlash between joints, we achieve structures with reconfigurability through dilation factor ($\alpha(x,y)$) variation. (B) Unit cells are rotating square auxetic mechanisms. (C) By adding actuators with position feedback, we create a robot that can manipulate an object on a surface.}
    \label{fig:figure1}
\end{figure}

\newpage

Demonstrating new capabilities in real-time, 3D reconfiguration of auxetic structures for multi-functional and adaptive surface manipulation is novel and valuable. If we are able to adjust the dilation factor ($\alpha(x,y)$) for cells across an auxetic lattice, we may generate multiple 3D shapes from one reconfigurable lattice structure. We have accomplished this and our demonstration is shown in Figure \ref{fig:figure1}. Methods in prior work produced a range of expansion over an auxetic lattice to enable improvements in surface conformity while maintaining structural integrity \cite{MackertichSengerdy_reconfigelectromag_2023, bertoldi_auxetic_2010}. From our work, we observe that current auxetic fabrication processes produce structures at a limited range of adaptability and responsiveness compared to what is needed for dynamic robotic applications \cite{Holmes_shapeshift_20190118, kristensen2016flexible_ep}. Smooth transitions of expansion and stiffness ensure that structures achieve designed deformation patterns according to surface geometry requirements. We believe this may be achieved by Reconfigurable Auxetic Devices (RADs) that derive their compliance from a combination of elastic rod theory and backlash coupling between unit cells.

\begin{figure}[!h]
    \centering
    \includegraphics[width=4in]{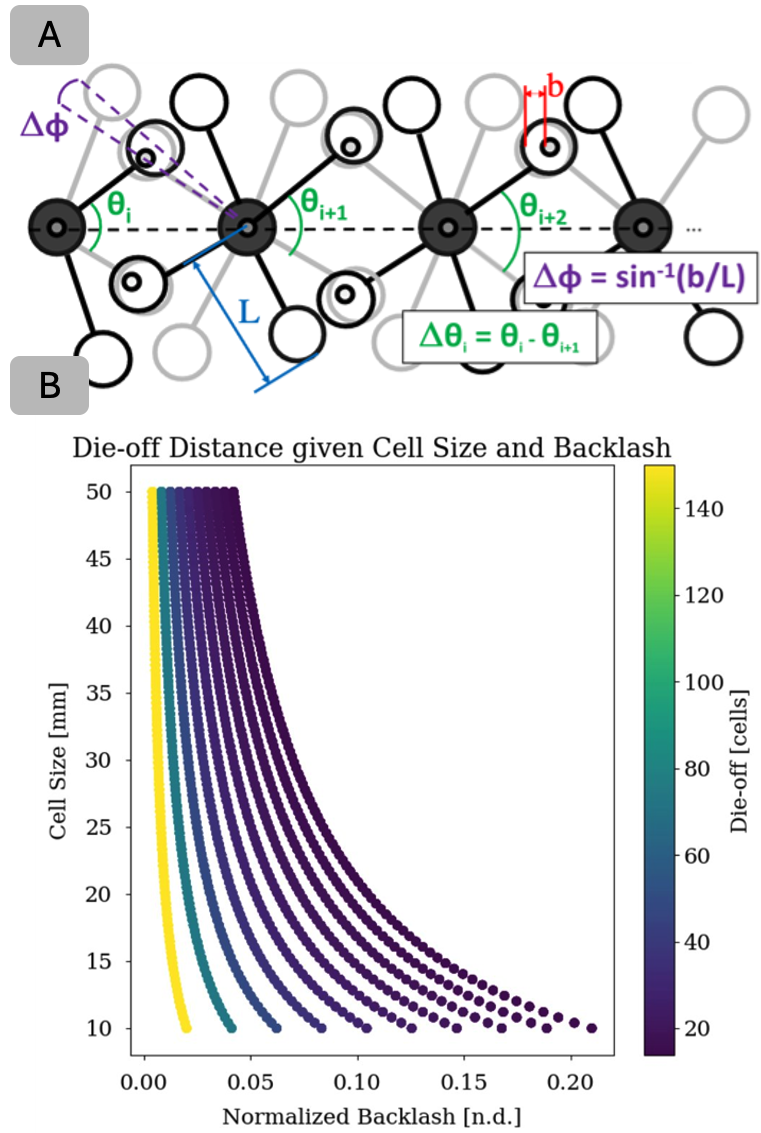}
    \caption{(A) A simplified diagram of a chain of RADs unit cells. (B) Relationship between normalized backlash, cell size, and the distance of intercell coupling ('die-off').}
    \label{fig:figure2}
\end{figure}

We see RADs as a tool for rapid production of custom surface shapes that achieve specific surface conformity under different loading conditions. The lattice unit cell is an auxetic mechanism produced with a specific amount of backlash between cells and may be built with desired variable expansion ratios and stiffness gradients across a wide range. Our method invokes soft compliance via this backlash. Figure \ref{fig:figure2} shows how backlash allows for variation in unit cell dilation up to a finite distance. In our work, the unit cell geometry was varied across multiple configurations. Each lattice we built was modeled in software that we also used to control the system. To test the prototypes, we integrated optical tracking, comparing a desired shape to lattice surface measurements to validate the method. 

In this paper, we:
\begin{itemize}
    \item Establish a design-to-fabrication method for reconfigurable, robotic surfaces using auxetic mechanisms
    \item Show that our RADs method enables geometrically complex, variable Poisson's ratio zones in auxetic structures  
    \item Demonstrate that the compliance of mechanical metamaterials may be engineered by specifying the degree of backlash in joints
    \item Show the utility of RADs, by producing dynamic manipulation surfaces from a single, interconnected lattice structure
\end{itemize}

\section{Background}
\thispagestyle{empty}
Reconfigurable auxetic structures for robotic surface manipulation provide adaptive interfaces that conform to various surface geometries and contact requirements. In these systems, prior work has shown that auxetic, kirigami, and origami structures are useful for applications ranging from manufacturing to robotics, where systems need to adapt to a wide range of surface conditions \cite{kim2013soft, bertoldi_auxetic_2010, Sempuku_kirigami_2021}. In prior work, authors utilized auxetic structures with varying levels of negative Poisson's ratios and expansion characteristics \cite{Lipson2000} and constructed manipulation interfaces to adapt to diverse surface geometries \cite{Liu_roboticsurface_2021, gonzalez2023constraint}. These works show a range of auxetic materials with variable expansion properties for adaptive surface contact. However, many prior fabrication methods require specialized tooling or complex assembly processes and focus on static configurations and 1D transformations instead of building reconfigurable structures. Today, new computational modeling methods combined with embedded actuation enable us to explore unit cell geometry for reconfigurable surface shape changes. This capability is functionally enabled by spatially variable auxetic properties.

In difference with prior work, an underactuated lattice control system may coordinate multiple regions of dilation through integrated servo encoder feedback to adjust surface shape. This control approach is reminiscent of adaptive biological interfaces, where muscles cells actuate in serial and parallel arrangements \cite{laschi2014octopus}. 2D conformity responses of flat lattices of RADs are shown in Figure \ref{fig:figure3}. Previous work mapped input geometry to surface shape characteristics and was locally-programmed, in contrast with variable conformity devices with selective surface adaptation. Notably, prior work on reconfigurable auxetics did not explore the boundaries of real-time dilation variation across devices for use in dynamic surface manipulation. We explored this concept through RADs unit cell design.

\section{RADs Cell Design}
We modeled the RADs cell after the auxetic rotating squares pattern. Figure \ref{fig:figure2} depicts the key geometry parameters in our model and the 'die-off distance', which is the range of intercell coupling given backlash between cells. Normalized backlash is given by Equation \ref{eq:normalized_backlash}. The coupling of adjacent cells is estimated by using a Realized Linear Unit (ReLU) function, shown in Equation \ref{eq:relu} and the relationship between the theta parameter in Figure \ref{fig:figure2} and dilation factor $\alpha$ is given by Equation \ref{eq:alpha_theta}. This set of simple relationships allows us to estimate cell coupling based on the backlash geometry in cell joints.

\begin{figure}[!h]
    \centering
    \includegraphics[width=4in]{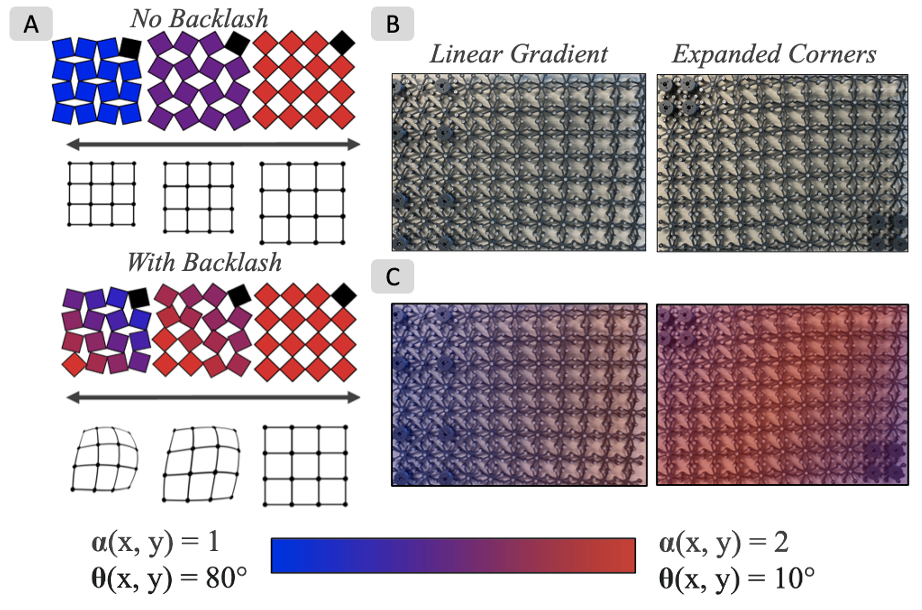}
    \caption{(A) Cartoon of key terms in simplified cell coupling model. (B) The range of intercell coupling (die-off distance), is a function of cell geometry when backlash is included.}
    \label{fig:figure3}
\end{figure}

\begin{figure}[!h]
    \centering
    \includegraphics[width=5in]{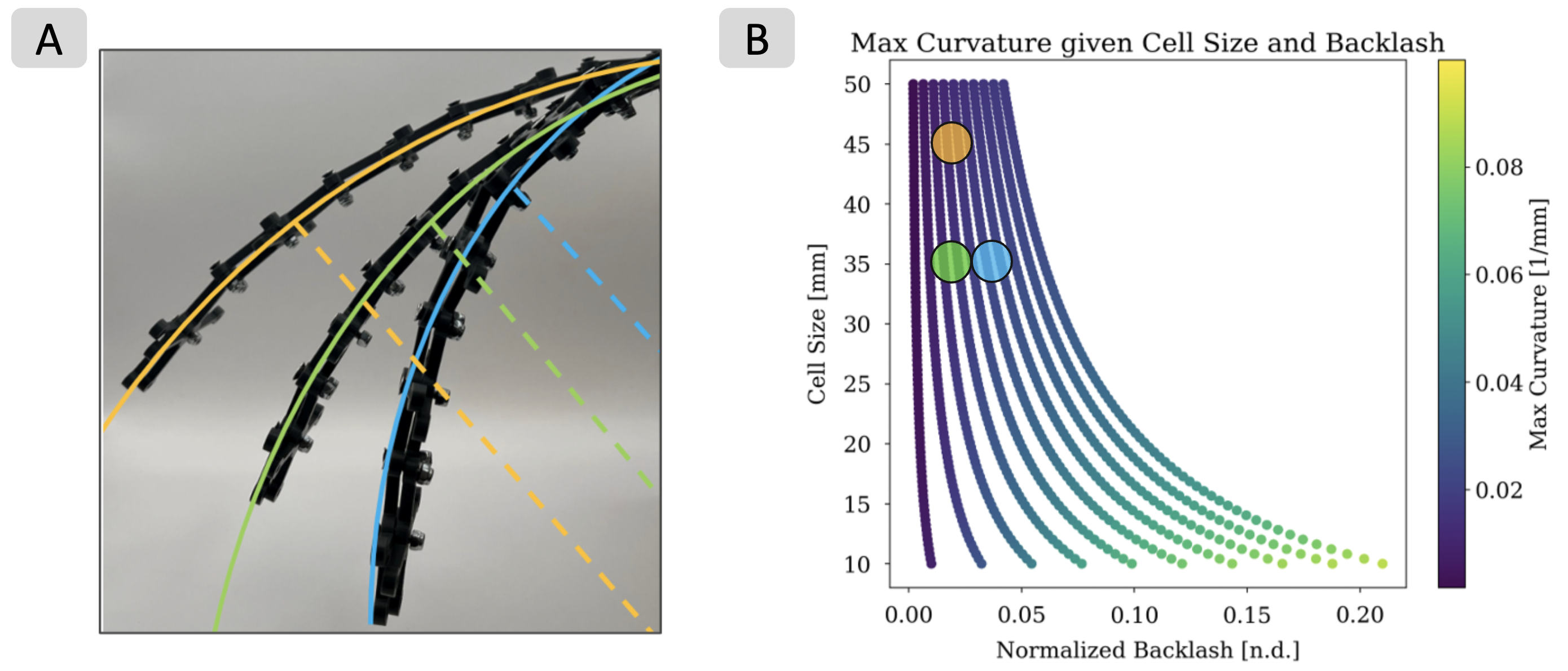}
    \caption{(A) Curvature of three auxetic linkages varying in accordance with different arm length and normalized backlash. (B) The points shown align with model results.}
    \label{fig:figure_curvature}
\end{figure}

The surface adaptation of RADs lattices was achieved by employing variable auxetic behavior across the manipulation interface via backlash between cells. The device structure enables multi-directional surface conformity while maintaining manipulation capability through a soft rubber skin placed on the lattice surface. This results in a device with non-permeable barrier and surface interpolating membrane. We determined the properties of the rubber skin by identifying a parameter set that would be stretchable without tearing at the strains induced by RADs motion. This gives us a non-permeable, manipulation device capable of conforming to adaptive surfaces. Figure \ref{fig:figure3} shows the structural shape variation due to backlash between cells. Figure \ref{fig:figure_curvature} shows the relationship we measured between cell geometry and linkage curvature limits.

\begin{equation} \label{eq:normalized_backlash}
    b_{norm} = \frac{b}{L}
\end{equation}

\begin{equation}\label{eq:relu}
    f(x) = max(0,x-b)+min(x+b,0)
\end{equation}

\begin{equation}\label{eq:alpha_theta}
    \theta [degrees] = 70 \alpha - 60 
\end{equation}

\section{RADs Lattice Fabrication}
\begin{figure}[!h]
    \centering
    \includegraphics[width=5in]{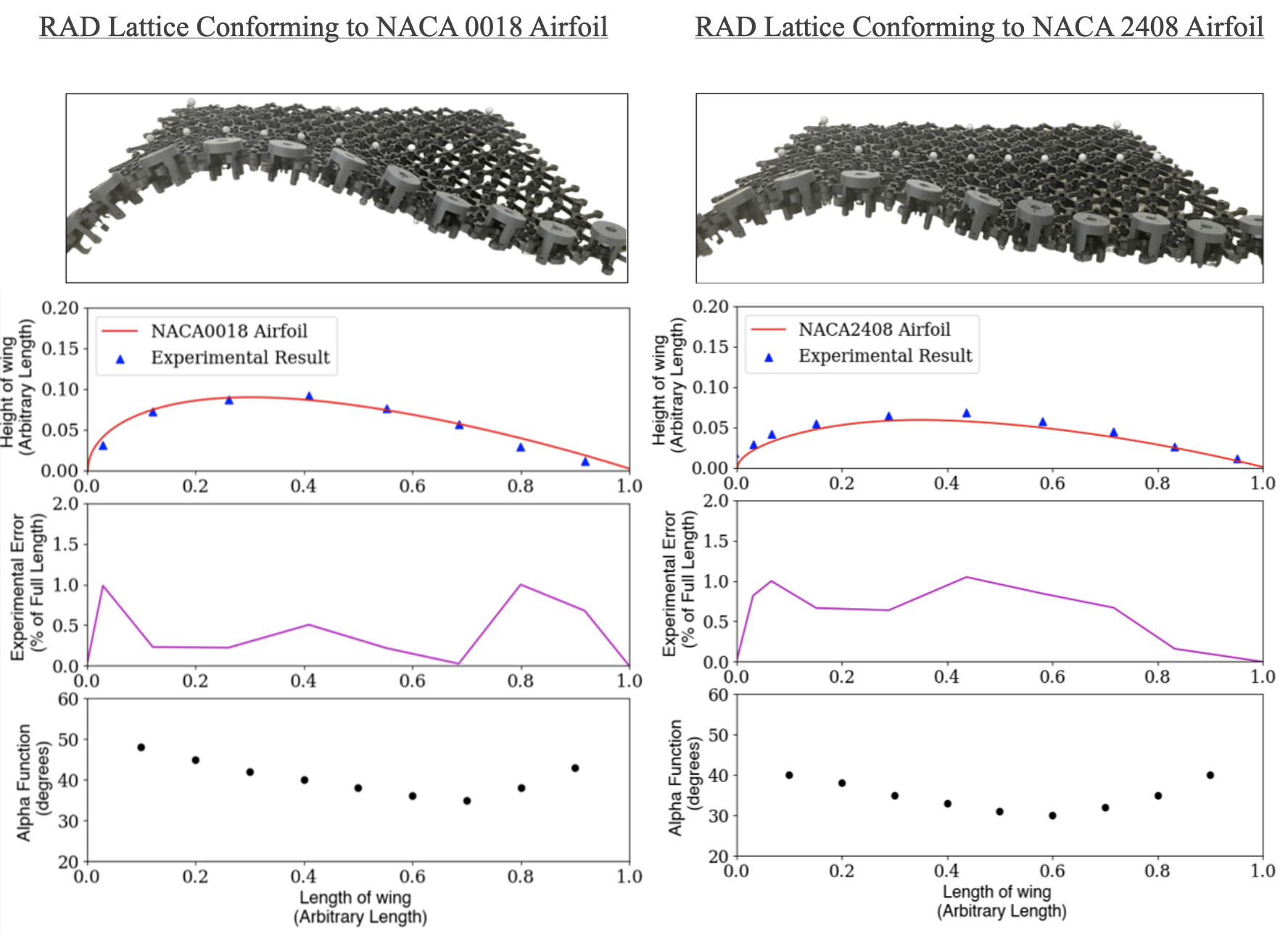}
    \caption{(Left) A RADs lattice with 11x8 layout and mechanical locks conforming to a NACA 0018 airfoil profile. (Right) The same lattice with alternative locks conforming to the NACA 2408 profile.}
    \label{fig:figure_airfoillattice}
\end{figure}

To fabricate RADs lattices, we first developed a parametric CAD model of the rotating-square auxetic unit cell. The cell consists of two concentric parts with four joints each. The geometry of the cell was set such that it maintained a negative Poisson’s ratio of -0.4 and a normalized backlash value of $b = 0.1$ while also providing structural integrity to allow deformation with the actuation range. The cell was scaled to a 35 mm side length so that prototype 8x11 and 11×15 unit lattices were portable. A set of tests to conform the 8x11 prototype are shown in Figure \ref{fig:figure_airfoillattice}. These fixed locks set the dilation factor of select cells such that the lattice conformed to an National Advisory Committee for Aeronautics (NACA) airfoil shape once a boundary condition was placed on the sides of the lattice.

The components of the lattice were made using Prusa MK4 3D printer with PLA filament, chosen for its reliability, ease of processing, and consistent fabrication of holes within a 0.1 mm tolerance. After the printing process all parts were processed to remove residual filament for precise alignment of the joints and hinges in order to ensure consistent backlash throughout the structure's joints. The cells were then concatenated into a continuous grid. Servo actuators were integrated within specific lattice members in a pattern formation on the underside of the lattice. Mounting the servos beneath the structure preserved the upper surface of the structure allowing for no obstruction when deforming and wire management with routing to breadboard and power supply below the system. The configuration chosen reduced mechanical interference, minimized visual clutter, and facilitated modular reprogramming of the actuation sequences. 

The lattice assembly was suspended beneath a clear plastic sheet that supports the lattice, allowing the structure to freely deform by product of its own weight while also being elevated. During actuation, gravity influences the lattice causing it to drape downward, with a neutral curvature across the surface determined by the boundary conditions and cell geometry. This state represents the neutral equilibrium configuration. Through integrated actuation, the suspended lattice shape and curvature may be dynamically tuned to produce continuous deformations that conform into unique, alternative surfaces within its configuration space. Electrical connections were routed below the structure, and the servo network was programmed to allow group control through the residual deformation of cells within a specified die-off distance of the servo actuated cell, enabling both localized and global shape transformations. The wiring layout and actuator placement strategy were informed by our model, which provided estimates of the system workspace given control density and deformation sensitivity.

The lattices were tested under different actuation sequences to validate its range of motion and force response. The interplay between gravity-induced deformation and servo-driven correction provided a practical mechanism for adaptive shape control within the structure’s configuration space, confirming the functional viability of the RAD lattice as a reconfigurable, auxetic-based surface manipulation system. 

\begin{figure}[!h]
    \centering
    \includegraphics[width=3.3in]{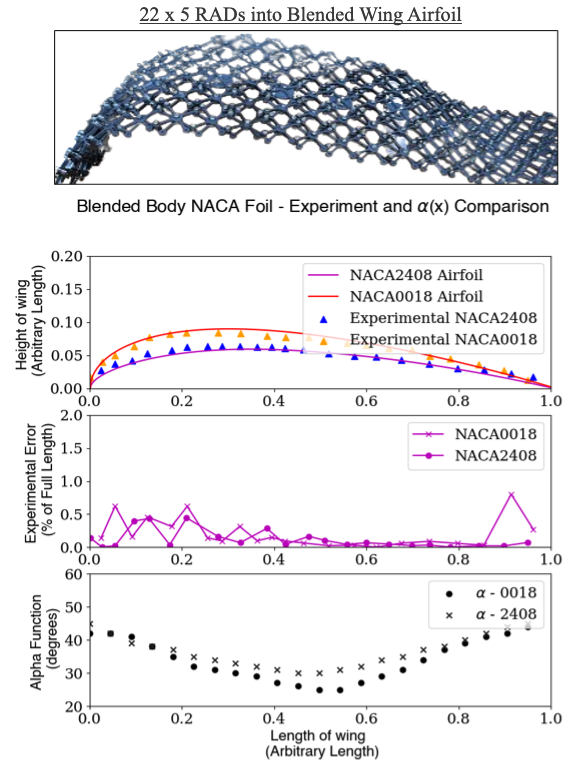}
    \caption{A RADs lattice using mechanical locks may be set into a complex curved contour, such as two NACA airfoil profiles.}
    \label{fig:figure4}
\end{figure}

\section{Modeling Backlash to Morphing Surfaces}
We produced software to model the deformation of RADs lattices within an arbitrary 3D system geometry. The model assumes that different auxetic regions are locally isotropic with smooth transitions in properties between regions. By assigning regions to varying Poisson's ratios and elastic moduli, we simulated how devices respond to pneumatic pressures and surface contact forces. The applied pressure in the model is distributed through the embedded pneumatic network with realistic boundary conditions matching the physical RADs devices. Figure \ref{fig:figure4} 

Validation of the airfoil model to experimental results demonstrates that lattices of RADs. This validation holds for the elastic regime and controlled torques, with the model using nonlinear material properties to capture large deformation behavior.

\begin{equation} \label{eq:RoR}
    \theta_i = \theta_{i+1} \pm \Delta \phi
\end{equation}
\thispagestyle{empty}

\section{Prototype Results}
The servo actuation system consisted of motors connected to specific unit cells throughout the auxetic lattice. We calibrated servo responses at intervals of 0.1 radians. Surface manipulation was driven by coordinated control with feedback from integrated force sensors for contact detection. The control system used regulation to achieve desired surface conformity. Once set, measurements of surface area and conformity were obtained using photography. The surface characteristics were calculated based on total contact area across the lattice. The system was able to increase or decrease surface area to match target surface geometry as external conditions changed. The surface conformity of devices was measured using 3D scanning for different conditions and compared against predicted surfaces from other models. The positions and contact patterns of devices are shown in Figure \ref{fig:figure7}.

\begin{figure}
    \centering
    \includegraphics[width=3.3in]{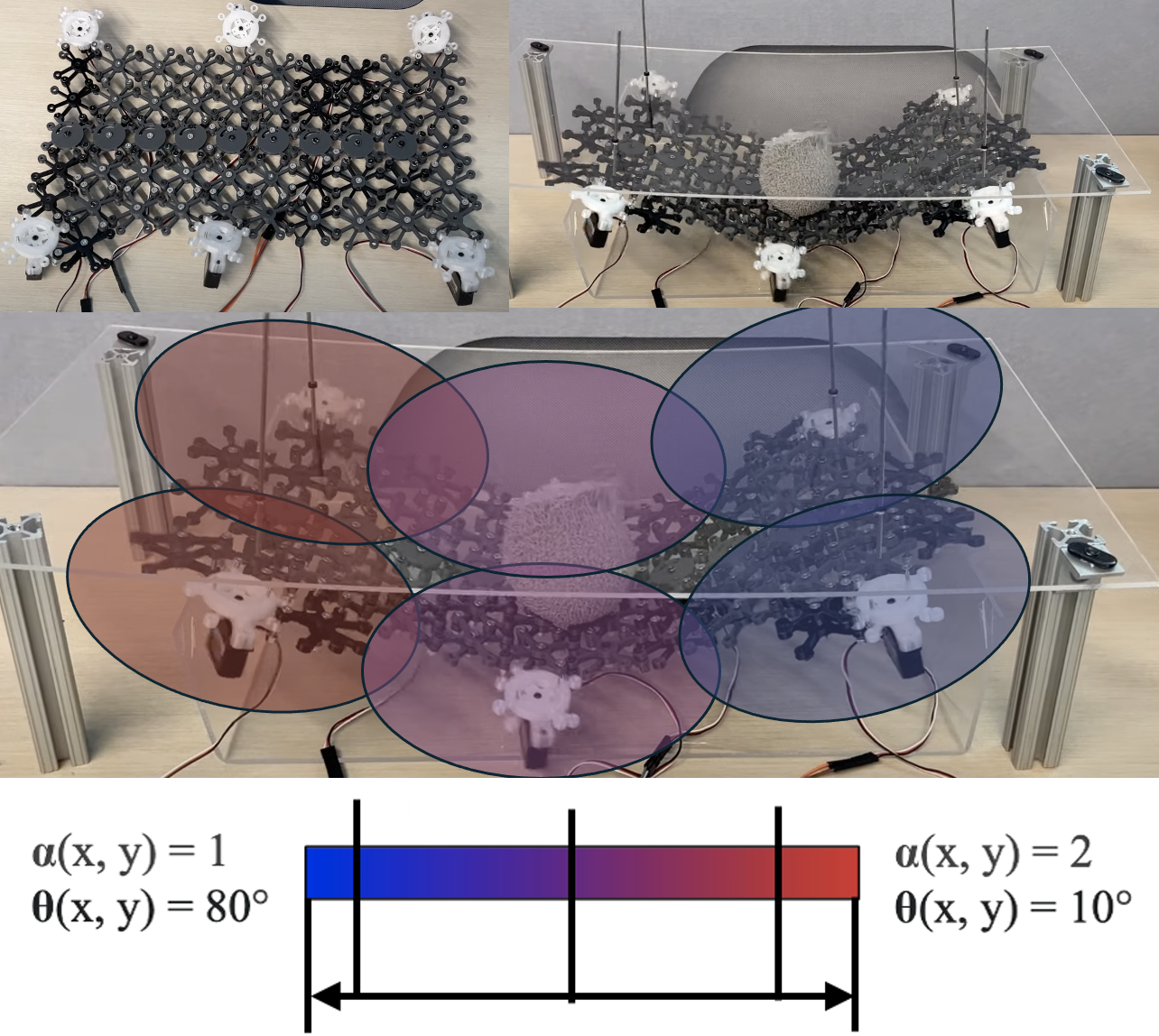}
    \caption{RADs prototype with six servos. (Top left) Lattice structure without frame. (Top right) Auxetic surface robot on frame. (Bottom) Side view showing six regions of expansion and contraction varying across the dilation spectrum.}
    \label{fig:figure5}
\end{figure}

\subsection{Validation}
\thispagestyle{empty}
The motion of a foam ball on the 11x15 RADs prototype was measured with an optical tracking system and compared against the predicted motion from our model based on the ReLU intercell coupling assumption. The ball's motion from experimentation and modeling demonstrate notably different behaviors. We noted discrete stages of interaction between the RADs surface and the ball. When the rows of servos on the RADs lattice were triggered in the right order and frequency, the static friction and inertia of the ball was overcome and back and forth motion was achieved. When the order and frequency are adjusted, there are regimes where the ball is static at different equilibrium points on the surface, or even rolls back against the servo action.

The testing results demonstrated less than 12\% difference between modeled and measured surface area for the rotating squares unit cell geometry chosen.

\begin{figure}[!h]
    \centering
    \includegraphics[width=3.5in]{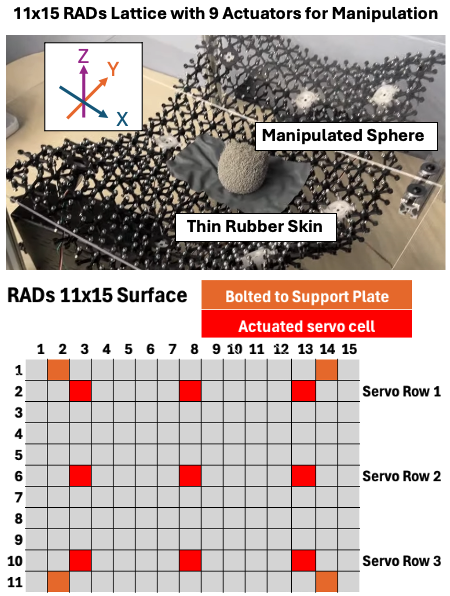}
    \caption{Experimental setup of 11x15 RADs lattice with 9 servos in three rows.}
    \label{fig:figure6}
\end{figure}

\begin{figure*}[!h]
    \centering
    \includegraphics[width=7in]{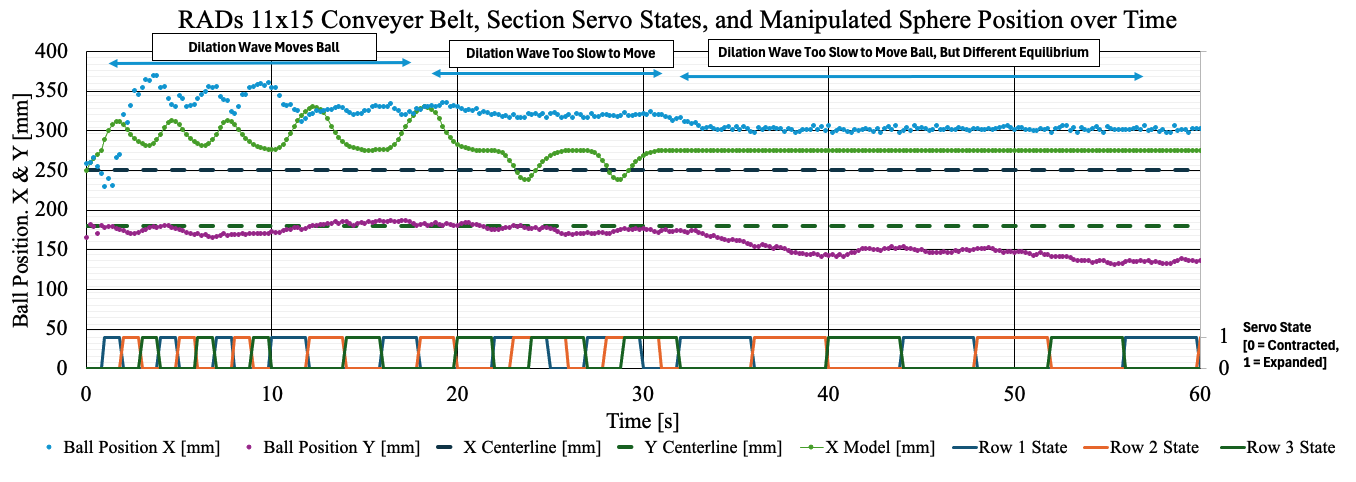}
    \caption{Data from experimental RADs lattice robot surface shown in Figure 6 compared with model results based on activation function intercell coupling.}
    \label{fig:figure7}
\end{figure*}

\thispagestyle{empty}

\section{Discussion}
The lattices demonstrated in this work derive their compliance and shape-morphing from backlash between mechanical joints. This method allows for stiffer base materials to generate curved structures, similar to how chainmail is stiff in bulk properties yet compliant within a configuration space \cite{wang2021structured}. Fabricating soft surfaces with these latticed RADs offers benefits relative to conventional surface manipulation methods. RADs enables a production method for fabricating customized complex structures with spatially tunable mechanical properties. By allowing control over the shape, this method may also enhance the material elasticity and load-bearing capacity of substrate auxetic surfaces for additive manufacturing \cite{muth2014embedded}. The use of RADs manipulation underscores the potential of this method to provide a versatile and effective approach to advancing new surface manipulation capabilities for soft robotics. Alternative actuation methods, such as shape memory alloys or electromagnetic systems, may be used for applications requiring different response characteristics \cite{kotikian20183d}. This technology provides an opportunity to overcome limitations associated with previous use of static auxetic materials in robotic surface manipulation systems . The most conformable regions in biological manipulators are the contact surfaces that directly interface with target objects. These surfaces, featuring hierarchical structures and variable stiffness, are critical for achieving optimal contact and force distribution. The contact regions are located at the interface between the manipulation system and target surfaces, enabling adaptive conformity and manipulation. 
\thispagestyle{empty}

\section{Conclusions}
Through the development and testing of the RADs robotic surface, we demonstrated a framework for advancing soft surface control. The experimental validation of a RADs surface provides confidence in this methodology. Ongoing refinements in mechanical control and software design are expected to improve usability and functionality of this framework. Continued efforts will demonstrate new capabilities in underactuated control of mechanical metamaterials to rapidly generate target geometries, ready for integration into robotic systems. Next steps for work on RADs will involve increasing the technical complexity to produce convex bodies. Future work includes a plan to fabricate grippers with RADs to better generate complex and dynamic motion while applying force to an external body. This effort will be useful in demonstrating examples of utility. Additionally, the team will explore the incorporation of multi-functional materials in a RADs system to improve functionality.

Parallel to these technical improvements, real-world testing of manipulation and manufacturing surfaces produced with RADs will be integral to the method's evolution. Collaboration with soft robotics will hopefully provide insights into the practical usability and effectiveness of these reconfigurable surfaces to conduct valuable work or produce custom moulds rapidly. The iterative process of testing, feedback, and refinement will ensure that future RADs based structures are functional and user-friendly, contributing to greater utility for users of customized tooling processes.

\thispagestyle{empty}
\bibliographystyle{plain}
\bibliography{biblio.bbl}

\section*{Acknowledgment}
The authors acknowledge support from Northeastern University's Institute for Experimental Robotics (IER) and the colleagues of the Transformative Robotics Lab (TRL).

\section*{Conflict of Interest}
The authors declare no conflict of interest.

\section*{Data Availability Statement}
The data that support the findings of this study are available from the corresponding author upon request. GitHub repo with data and code at: https://github.com/TransformativeRoboticsLab/Auxetic-Backlash-Model

\section*{Funding Statement}
This work was supported by Northeastern University's Institute for Experimental Robotics (IER).

\section*{Ethics Approval Statement}
Not applicable.

\section*{Author Contributions}
JNM collected all data and 2D modeling results.

\noindent AM provided fabrication and testing of structures.

\noindent AI provided 3D modeling results.

\noindent JIL provided advice and writing support.

\thispagestyle{empty}

\end{document}